\documentclass[conference]{IEEEtran}
\IEEEoverridecommandlockouts
\usepackage[belowskip=-10pt,aboveskip=0pt]{caption}
\usepackage{cite}
\usepackage{amsmath,amssymb,amsfonts}
\usepackage{algorithm,algorithmic}
\usepackage{graphicx}
\usepackage{subfig}
\usepackage{float}
\usepackage{textcomp}
\usepackage{xcolor}
\usepackage{bm}
\def\BibTeX{{\rm B\kern-.05em{\sc i\kern-.025em b}\kern-.08em
    T\kern-.1667em\lower.7ex\hbox{E}\kern-.125emX}}
    
\newcommand{\deltamethod}{delta-method}
\begin{document}

\title{SOLBP: Second-Order Loopy Belief Propagation for Inference in Uncertain Bayesian Networks
\thanks{This research was partially supported by the Army Research Office under contract W911NF-15-1-0479 and partially supported by the U.S. Army Research Laboratory and the U.K. Ministry of Defence under agreement W911NF-16-3-0001. The views and conclusions contained in this document are those of the authors and should not be interpreted as representing the official policies, either expressed or implied, of the U.S. Army Research Laboratory, the U.S. Government, the U.K. Ministry of Defence or the U.K. Government. The U.S. and U.K. Governments are authorized to reproduce and distribute reprints for Government purposes notwithstanding any copyright notation hereon.}
}

\author{\IEEEauthorblockN{Conrad D. Hougen\IEEEauthorrefmark{1}, 
Lance M. Kaplan\IEEEauthorrefmark{2}, 
Magdalena Ivanovska\IEEEauthorrefmark{3}, 
Federico Cerutti\IEEEauthorrefmark{4},\\
Kumar Vijay Mishra\IEEEauthorrefmark{2} and 
Alfred O. Hero III\IEEEauthorrefmark{1}}\\
\IEEEauthorblockA{\IEEEauthorrefmark{1}\textit{University of Michigan}, Ann Arbor, MI, USA \\
Email: chougen@umich.edu, hero@umich.edu}
\IEEEauthorblockA{\IEEEauthorrefmark{2}
\textit{DEVCOM Army Research Laboratory}, Adelphi, MD, USA \\
Email: lance.m.kaplan.civ@army.mil, vizziee@gmail.com}
\IEEEauthorblockA{\IEEEauthorrefmark{3}
\textit{BI Norwegian Business School}, Oslo, Norway \\
Email: magdalena.ivanovska@bi.no}
\IEEEauthorblockA{\IEEEauthorrefmark{4}\textit{University of Brescia}, Brescia, Italy \\
Email: federico.cerutti@unibs.it}}


\maketitle

\begin{abstract}
In second-order uncertain Bayesian networks, the conditional probabilities are only known within distributions, i.e., probabilities over probabilities. The \deltamethod\ has been applied to extend exact first-order inference methods to propagate both means and variances through sum-product networks derived from Bayesian networks, thereby characterizing epistemic uncertainty, or the uncertainty in the model itself. Alternatively, second-order belief propagation has been demonstrated for polytrees but not for general directed acyclic graph structures. In this work, we extend Loopy Belief Propagation to the setting of second-order Bayesian networks, giving rise to Second-Order Loopy Belief Propagation (SOLBP). For second-order Bayesian networks, SOLBP generates inferences consistent with those generated by sum-product networks, while being more computationally efficient and scalable.
\end{abstract}

\section{Introduction}
Probabilistic graphical models (PGMs), such as Bayesian networks (BNs) and Markov random fields, encode conditional dependencies between variables of interest. In particular, BNs \cite{pearl.86} (Section \ref{sec:BN}) encode dependencies between variables using directed acyclic graphs (DAGs), parameterized by conditional probabilities. BNs are often used to infer the probability of values of unobserved variables, given values from a set of evidence variables. Standard BNs only consider \textit{first-order uncertainty}, where the network conditional probabilities are assumed to be known precisely. In these BNs, first-order inference methods are employed to predict the conditional probabilities of the network as point values.
 
Conversely, uncertain BNs (Section \ref{sec:UBN}) are BNs where the conditional probabilities are not known, or are known only up to intervals or distributions. The type of uncertainty in BNs can take several forms, leading to different categories of uncertain BNs, such as credal networks \cite{credal98}, valuation-based systems \cite{shenoy.89}, subjective BNs \cite{ivanovska.15}, and second-order BNs \cite{kaplan2018efficient}. A comprehensive survey of learning and reasoning for uncertain BNs is provided in \cite{rohmer2020uncertainties}.

Second-order BNs \cite{kaplan2018efficient}, in particular, consider \textit{second-order probabilities} where knowledge of the conditional probabilities is encoded by posterior distributions, which here are distributions over probabilities. The goal of second-order inference is to capture epistemic uncertainty in the model. Not only do we wish to make accurate inferences: we would also like to quantify our confidence level in the inferred values. Exact second-order inference for second-order BNs is intractable. Recent works \cite{cerutti2022handling,kaplan.ivanovska.TBD} have promoted methods for second-order inference which involve translating them into sum-product networks (SPNs) (Section \ref{sec:SPN}) and then using the \textit{\deltamethod} (Section \ref{sec:delta}) to {\em approximate} the variance of the queried probabilities by performing a first-order Taylor expansion~\cite{kleiter.96,VanAllen.08}. The propagation of the means and variances by extending exact first-order inference methods via the \deltamethod\ leads to good approximations of the distributions of the queried probabilities \cite{kaplan2018efficient,cerutti2022handling}. However, sum-product networks do not scale well as the size of the BN grows.
As an alternative to the delta-method approximation, Monte Carlo sampling of first-order SPN inference could more accurately estimate the queried distribution but at even more computational expense.

One of the most efficient inference mechanisms for operating on arbitrary BNs is a generalization of belief propagation (Section \ref{sec:BP}) to networks containing loops, the so-called \textit{loopy belief propagation} (LBP) \cite{murphy1999loopy}. Here, a loop is any multi-path between a given pair of nodes. LBP usually, but not necessarily, converges to the correct inference \cite{ihler.05}. 

Our primary contribution, Section \ref{sec:solbp}, is an extension of LBP to the setting of second-order BNs with discrete-valued variables, where the conditional probabilities are Dirichlet-distributed. We call the resulting method Second-Order Loopy Belief Propagation (SOLBP). 
Indeed, using SPNs with \deltamethod\ approximation will constrain the size of the BN that we wish to perform inference on. Techniques that are commonly used for second-order inference, such as belief propagation, have so far been limited to polytree networks, where the graph structures are disallowed from containing loops. The experimental results in Section~\ref{sec:evaluation} 
show that the inferences generated by SOLBP match those generated by the \deltamethod\ in SPNs when 
LPB correctly converges,
confirming correctness of SOLBP, at a lesser computational cost. Finally, we draw our conclusions in Section~\ref{sec:conclusions}.

\section{Background in Bayesian Networks}

\label{sec:BN}

Associated with any BN is a directed acyclic graph (DAG), where each node is a random variable, and links represent conditional dependencies \cite{jensen.07}. Each child variable, $Y$, with parents $\mathbf{X} = [X_1,...,X_{N_p}]$, has an associated conditional probability table $p(Y=y|X_1=x_1,...,X_{N_p}=x_{N_p})$. Here, we have assumed that the BN is composed of discrete-valued variables, which is the common case. A variable, $Y$, with no ancestors, is then associated with an unconditional probability table $p(Y)$. Each variable $Y$ has a finite set of mutually exclusive states, $\mathbb{Y}$.

A given graph with fixed structure---i.e., the set of nodes and edges---represents a family of BNs parameterized by the probabilities $\bm{\theta}$, which we interchangeably refer to as the \textit{parameters} or \textit{conditionals}. Each parameter corresponds to a particular entry in the conditional or unconditional probability tables of the BN.

The joint probability of variables $Y_1,...,Y_N$ is
\begin{equation}\label{eqn:jointprob}
    p(y_1,...,y_N; \bm{\theta})=\prod_{i=1}^{N} \theta_{y_i|\mathbf{x}_i},
\end{equation}
where $y_i\in \mathbb{Y}_i$ is the value that $Y_i$ is assigned, and $\mathbf{x}_i \in \prod_{j=1}^{N_{p,i}} \mathbb{X}_{i,j}$ represents a specific assignment of the parents of variable $Y$. For any assignment of the parents of $Y$, the conditionals, $\theta_{y_i|\mathbf{x}_i}$ form a distribution, i.e. 
\begin{equation}\label{eqn:secondconstraint}
    \sum_{y_i\in \mathbb{Y}_i} \theta_{y_i|\mathbf{x}_i} = 1, \forall \mathbf{x}_i\in \prod_j \mathbb{X}_{i,j}, \forall i.
\end{equation}

One of the main inference tasks in BNs is to determine the conditional probabilities of unobserved variables, given observations of evidence variables. We will focus here on the sub-task of computing the marginals of the nodes, i.e.~determining the probability distribution of each variable in the network $Y_i$ upon  observing the values of a set of variables $\mathbf{E}$, i.e.~computing  $p(y_i|\textbf{e})$, for each $y_i\in \mathbb{Y}_i$. We denote this distribution by $p(Y_i|\mathbf{e})$.
If $\mathbf{E} = \emptyset$, then $p(y_i|\textbf{e})=p(y_i)$. In general the computation of a queried probability for node $y_i$ requires calculations of the marginal
\begin{equation}
p(y_i,\mathbf{e}) = \sum_{\mathbf{y}\sim \mathbf{e} \cup \{y_i\}}\left (\prod^N_{j=1} \theta_{y_j|\mathbf{x}_j}\right)\;,
\end{equation}
where ``$\sim$'' is used to denote compatibility of variable assignments. 
By the rule of conditional probability, 
\begin{equation} \label{e:query}
p(y_i|\mathbf{e}) = \frac{p(y_i,\mathbf{e})}{\sum_{y_i \in \mathbb{Y}_i} p(y_i,\mathbf{e})}.
\end{equation}
Note that the computation of the probability is a nonlinear function of the set of conditional probabilities $\Theta$ for the BN. The following subsections review methods to compute (\ref{e:query}).

\subsection{Belief Propagation}
\label{sec:BP}
Belief propagation, introduced in \cite{pearl.86}, uses message passing to efficiently compute exact inferences over polytrees. Even when the BN is not a tree, it may be possible to segment the BN into multiple tree structures, giving rise to the junction-tree algorithm \cite{madsen1999lazy}, which reduces to belief propagation for networks without loops. In belief propagation, each node in the BN both passes and receives messages to or from neighboring nodes. Specifically, each node receives $\lambda$-messages from its children and $\pi$-messages from its parents, which are then used to calculate outgoing $\lambda$ and $\pi$ messages.

The $\pi$ message from a parent node $X_1$ to a child node $Y$ encodes the probability of the parent variable, conditioned on the evidence coming from $X_1$. If the evidence includes a particular value for $Y$, then the internal $\bm{\pi}_Y$ value is a one-hot vector, i.e., $\pi_Y(y) = \delta_{y,e_Y}$ where $e_Y$ is the value of $Y$ in the evidence $\mathbf{e}$.  Otherwise, node $Y$ collects its $\pi$-messages from its parents $\mathbf{X} = [X_1,\ldots,X_{N_p}]$. Once node $Y$ receives these messages, the node $Y$ can compute its internal $\pi$-values:
\begin{equation} \label{fprop}
    \pi_Y(y) = \sum_{\bm{X}} 
    \theta_{y|x_1,\ldots,x_{N_p}} \prod_{i=1}^{N_p} \pi_{Y,X_i}(x_i).
\end{equation}

Similarly, the internal $\bm{\lambda}_Y$ value for node $Y$ is a one-hot vector if the evidence includes the value of $Y$, i.e., $\lambda_Y(y) = \delta_{y,e_y}$.  Otherwise once the $\lambda$-messages are received by node $Y$ from all of its children, $\mathbf{Z} = [Z_1,\ldots,Z_{N_c}]$, the internal value is accumulated as a product:
\begin{equation} \label{lambdafusion}
    \lambda_Y(y)=\prod_{j=1}^{N_c} \lambda_{Y,Z_j}(y).
\end{equation}

Once a node can compute its internal $\bm{\pi}_Y$ and $\bm{\lambda}_Y$ values, it can compute its value probabilities as
\begin{equation} \label{finalfusion}
    p(y|\mathbf{e})=\dfrac{\lambda_Y(y)\pi_Y(y)}{\sum_{y'\in\mathbb{Y}}\lambda_Y(y')\pi_Y(y')}.
\end{equation}

A node can only send out a message to a neighbor once it has received its messages from all other neighbors. The node sends out a $\pi$-message to a child node $Z_j$ once it has computed its internal $\bm{\pi}_Y$ value and received $\lambda$-messages from all the other children in $\mathbf{Z}$.  This message is computed as
\begin{equation} \label{pifusion}
    \pi_{Z_i,Y}(y) = \pi_Y(y)\prod_{j\neq i} \lambda_{Y,Z_j}(y),
\end{equation}
where $\bm{\lambda}_{Y,Z_j}$ is the $\lambda$-message that child $Z_j$ sent to node $Y$. 
Likewise, node $Y$ can send out a $\lambda$ message to a parent node $X_i$ once it has computed its internal $\bm{\lambda}_Y$ value and received $\pi$-messages from all the other parents in $\mathbf{X}$.  The message is computed as
\begin{equation}\label{backprop}
    \lambda_{X_i,Y}(x_i)=\sum_{y\in \mathbb{Y}}\lambda_Y(y)\sum_{\bm{X}\sim x_i}\theta_{y|x_1,\ldots,x_{N_p}}\prod_{j\neq i}\pi_{Y,X_i}(x_j),
\end{equation}
where $\bm{\pi}_{Y,X_i}$ is the $\pi$-message that parent $X_i$ sent to node $Y$.

Initially, only nodes that do not have parents can compute their internal $\bm{\pi}_Y$ values where $(\ref{fprop})$ simplifies to $\pi_Y(y) = \theta_y$. Likewise, nodes that have no children can assign their internal $\bm{\lambda}_Y$ values to the non-informative identity value for product fusion, i.e., $\lambda_Y(y) = 1$. 



Nodes which do not have parents do not send $\lambda$-messages, and similarly, nodes without children do not send $\pi$-messages. Once all of the nodes in the network have sent their respective $\lambda$ and $\pi$ messages, each node $Y$ will have an associated belief, $p(y|\mathbf{e})$, given the instantiated variables. The instantiated variables are also called evidence, which we have denoted by the boldface $\mathbf{e}$, where we use the boldface notation to indicate that the evidence is a vector of multiple variables.

Belief propagation is designed to provide exact inference for polytree BNs. It can be shown that for polytree networks, all nodes will be able to send their $\pi$- and $\lambda-$message to each other, and the probabilities computed by (\ref{finalfusion}) are equivalent to (\ref{e:query}). 
An extension -- Loopy Belief Propagation \cite{murphy1999loopy} -- is based on the idea to run Belief Propagation on a graph containing loops using a fixed-point iteration procedure, despite the fact that the presence of loops does not guarantee convergence. As our contribution directly expands on LBP, we will illustrate its details in Section \ref{sec:solbp} in the context of our proposal.


\subsection{Sum-Product Networks for Inference}
\label{sec:SPN}

Another approach to inference in BNs takes advantage of a known technique for translating BNs into Sum-Product Networks (SPNs) as pioneered by Darwiche \cite{darwiche}. An SPN consists of a rooted DAG of interior operator nodes (e.g. sum or product operations) and leaf nodes which are associated with indicator variables. In order to convert a BN into an SPN, the basic idea is to introduce indicator variables $\lambda_{y_i};\ i=1,\ldots,n$ such that
\begin{equation}
    \lambda_{y_i} = \begin{cases}
      1, & \text{if}\ Y_i = y_i \\
      0, & \text{otherwise}
    \end{cases}.
\end{equation}
When combined with the BN parameter variables $\bm{\theta}$, we may write down a canonical polynomial for any particular BN. Given the canonical polynomial, it is relatively straightforward to form an associated SPN graphical model, by means of variable elimination. The specific SPN structure is not unique and will depend on the order in which variables are eliminated.

The indicator variables of the leaf nodes in an SPN represent the assignment states of variables from the original BN from which the SPN was derived. By setting the indicator nodes of the SPN to either 1 or 0, we can efficiently compute $p(\mathbf{e})$, where $\mathbf{e}$ represents the evidence, i.e. the values of the observed variables. In order to accomplish this, we set the values of the leaf indicators to 0 if the conditional probabilities associated with each leaf are inconsistent with the evidence. A forward pass from the leaf nodes to the root of the SPN performs marginalization of (\ref{eqn:jointprob}) over all variables which are unobserved:
\begin{equation}
    p(\mathbf{e})=\sum_{\mathbf{y}\sim \mathbf{e}} \prod_{i=1}^N \theta_{y_i|\mathbf{x}_i}.
\end{equation}
An SPN readily performs this operation during a single forward pass.

Additionally, a backward pass through an SPN can be used to efficiently compute partial derivatives of the evidence likelihood with respect to each of the BN parameters, i.e. partial derivatives of the form
$$\frac{\partial p(\mathbf{e};\bm{\theta})}{\partial \theta_{y_k|{\mathbf{x}_k}}}.$$
These partial derivatives are beneficial for computing joint probabilities $p(y_k,\mathbf{e})$, for each of the variables $y_k$. These values are the outputs of the leaf nodes of the SPN:

\begin{equation}
    p(y_k,\mathbf{e})=\sum_{\mathbf{x}_{k}\in \prod_j \mathbb{X}_{k,j}} \frac{\partial p(\mathbf{e};\bm{\theta})}{\partial \theta_{y_k|{\mathbf{x}_k}}}\theta_{y_k|\mathbf{x}_k}.
\end{equation}

\section{Background in Uncertain Bayesian Networks}
\label{sec:UBN}

The conditional probabilities for BNs are either subjective in that they are provided by domain experts or they are learned from historical data. Here, historical data represents instantiations of the variable values sampled via the distribution described by the BN. As a result of learning with finite datasets, the conditional probabilities are not known precisely.  Nevertheless, the conditional probabilities are estimated from the data, usually via maximum likelihood estimation, and are treated as exact values during inference.

Second-order inference methods instead consider that knowledge of the conditional probabilities exhibit epistemic uncertainty encoded as the posterior distributions of the conditionals, given training data. As described in \cite{jensen.07}, when the dataset is complete in the sense that the values of all the variables are visible for each instantiation, the posterior distributions for the conditional probabilities factor as a product of Dirichlet distributions such that the posterior for $\bm{\theta}_{Y_i|\mathbf{x}_i}$ is
\begin{equation} \label{e:dirichlet}
f(\bm{\theta}_{Y_i|\mathbf{x}_i}) =  \frac{1}{B(\bm{\alpha}_{\theta_{Y_i|\mathbf{x}_i}})} \prod_{k=1}^{|\mathbb{Y}_i|}\left(\theta_{y_{i,k}|\mathbf{x}_i}\right)^{\alpha_{\theta_{Y_i|\mathbf{x}_i,k}}-1},
\end{equation}
where $k$ indexes the possible values of $Y_i$, $B(\cdot)$ is the multidimensional beta function
$$
B(\bm{\alpha}) = \frac{\prod_k \Gamma(\alpha_k)}{\Gamma(\sum_k \alpha_k)},
$$
and $\Gamma(\cdot)$ is the gamma function. The Dirichlet parameter $\alpha_k$ is one plus the number of times in the training data that node $Y_i$ took on a value of $y_{i,k}$ when the values of the nodes $\mathbf{X}_i$ took on values given by $\mathbf{x}_i$.  Remarkably, $\bm{\theta}_{Y_i|\mathbf{x}_i}$ is independent of all other conditionals $\bm{\theta}_{Y_i|\mathbf{x}'_i}$ and $\bm{\theta}_{Y_j|\mathbf{x}_i}$ for $\mathbf{x}_i \ne \mathbf{x}_j'$ and $i \ne j$.

In light of the posterior distributions for the conditional probabilities, second-order BN inference computes the distribution of the queried probabilities in (\ref{e:query}). Exact computation is intractable, but it can be approximated by leveraging the \deltamethod\ (see next subsection) that approximates the mean and covariance matrix for the queried values. To that end, we approximate the posteriors as Gaussian.  In light of the Dirichlet distribution given by (\ref{e:dirichlet}), the means for the values of $\bm{\theta}_{Y_i|\mathbf{x}_i}$ are given by
\begin{equation}
\mu[\theta_{y_{i,k}|\mathbf{x}_i}] = \frac{\alpha_{\theta_{Y_i|\mathbf{x}_i,k}}}{S_{Y_i|\mathbf{x}_i}}
\end{equation}
where the denominator
\begin{equation}
    S_{Y_i|\mathbf{x}_i} = \sum_{k=1}^{|\mathbb{Y}_i|}\alpha_{\theta_{Y_i|\mathbf{x}_i,k}}
\end{equation}
is known as the Dirichlet strength. 
The covariance for the terms in $\bm{\theta}_{Y_i|\mathbf{x}_i}$ is
\begin{equation}\label{e:fulllearncovariance}
\sigma[\theta_{y_i|\mathbf{x}_i},\theta_{y'_i|\mathbf{x}_i}] = \frac{\mu[\theta_{y_i|\mathbf{x}_i}](\delta_{y_i,y'_k}-\mu[\theta_{y'_i|\mathbf{x}_i}])}{S_{Y_i|\mathbf{x}_i}+1}\;.
\end{equation}

\subsection{The Delta-Method}
\label{sec:delta}

In general, reasoning in a BN can be viewed as passing the conditional probabilities through a nonlinear operation, e.g. (\ref{e:query}). The \deltamethod\ introduced in \cite{kleiter.96} approximates the mean and variance of the output of the nonlinear operation via a first-order Taylor series. 

The MeanVAR introduced in \cite{VanAllen.08} uses the \deltamethod\ to compute (\ref{e:query}) directly from the mean and variances of the conditional probabilities.  Alternatively, the second-order belief propagation (SOBP) in \cite{kaplan2018efficient} provides a more scalable solution by applying the \deltamethod\ to each step to compute the mean and covariances of the $\pi-$ and $\lambda-$messages (see (\ref{fprop})-(\ref{backprop})). Finally, \cite{cerutti2022handling,kaplan.ivanovska.TBD} apply the \deltamethod\ for each sum and product operation in the probabilistic circuit. 

In general, consider a set of $N$ input messages $\bm{\chi}_i = [\chi_i(x):x \in \mathbb{X}_i]$ for $i=1,\ldots,N$ with means $\mu[\bm{\chi}_i]$ and covariance $\sigma[\bm{\chi}_i]$ that are statistically independent of each other. Let the output message be related to the inputs via 
\begin{equation}
    \bm{\upsilon} =\mathbf{f}(\bm{\chi}_1,\ldots,\bm{\chi}_N),
\end{equation}
where $\mathbf{f} = [f_{\upsilon(u)}(\cdot):u\in \mathbb{U}].$ The \deltamethod\ linearizes the function $\mathbf{f}$ about the expected values of the input messages to approximate the output message as 
\begin{equation}
     \bm{\upsilon} \approx \mathbf{f}(\mu[\bm{\chi}_1],\ldots,\mu[\bm{\chi}_N]) + \sum_{i=1}^N \mathbf{J}_{\chi_i} (\bm{\chi}_i-\mu[\bm{\chi}_i]),
\end{equation}
where $\mathbf{J}_{\chi_i}$ is the Jacobian of the function $\mathbf{f}$ with elements representing the partial derivatives $\frac{\partial f_{\upsilon(u)}}{\partial \chi(x)}$ evaluated at 
the mean values of the $\bm{\chi}_i$ messages.

Because of the statistical independence between the messages, it follows that the mean and covariance matrices for the output message are
\begin{equation}\label{e:delta}
    \mu[\bm{\upsilon}]= \mathbf{f}(\mu[\bm{\chi}_1],\ldots,\mu[\bm{\chi}_N]), ~
    \sigma[\bm{\upsilon}] = \sum_{i=1}^N \mathbf{J}_{\chi_i} \sigma[\bm{\chi}_i] \mathbf{J}_{\chi_i}^T.
\end{equation}

The \deltamethod\ can extend an exact first-order BN inference engine into a second-order inference method that computes the means and covariances of the queried variables. In light of (\ref{e:delta}), the mean values are treated as the point probabilities in first-order inference. 

Any exact inference method can be extended to a second-order method via the delta-method.  For instance, second-order belief propagation was first developed in \cite{kaplan2018efficient}  for binary-valued polytree BNs.  Similarly, \cite{cerutti2022handling} provides the second-order extension of sum-product networks for general discrete-valued BNs.  Note that any exact first-order BN inference method is a composition of operations to transform the conditional probabilities into the inference.   
As argued in \cite{kaplan.ivanovska.TBD}, the variances computed by the second-order extensions of these inference methods through the delta-method are equivalent due to the chain rule of calculus.

\section{SOLBP Method}
\label{sec:solbp}

The main contribution of this paper is the extension of loopy belief propagation (LBP) for general BNs via the delta-method, which we refer to as SOLBP. This is motivated by the fact that belief propagation is more scalable than inference over sum-product networks. When LBP does converge to the exact inference, it represents a composition of operations leading to the exact inference function. Then, by chain-rule arguments in \cite{kaplan.ivanovska.TBD}, SOLBP should lead to the same variances as other exact second-order extensions.  

The basic concept is that all parent-to-child and child-to-parent messages are indexed. First, the messages are initialized to neutral values that are uninformative. A message is selected at random without replacement.  The mean and covariance of that message is updated based on the current values of its dependent messages.  Once the selection of messages have been exhausted, the round is completed. If the change in the mean values of the messages is small, the computation terminates.  Otherwise, another round is initiated. 
During a round, once all the parent or children input message have been updated, the node updates its internal $\pi-$ or $\lambda-$ messages, respectively. Algorithm~\ref{algA} summarizes the SOLBP steps.

\begin{algorithm}
\caption{Second-order loopy belief propagation.}
\label{algA}
\begin{algorithmic}[1]
\STATE \textbf{Input}: Evidence $\mathbf{e}$, $\mu[\theta_{Y|\mathbf{x}}]$ and $\sigma[\theta_{Y|\mathbf{x}}]$ for all $\theta_{Y|\mathbf{x}} \in \Theta$
\STATE \textbf{Output}: $\mu[\mathbf{p}(Y|\mathbf{e})]$ and $\sigma[\mathbf{p}(Y|\mathbf{e})]$ for all nodes $Y$ in the BN.
\STATE 
\STATE Set $\mathcal{M}^*$ to be the set of all possible inter-node messages
\STATE Set $\mathcal{M} \leftarrow \mathcal{M}^*$
\STATE For all $\bm{m} \in \mathcal{M}$, set $$\mu[\mathbf{m}] = \left \{ \begin{array}{cc}
\frac{1}{|\mathbb{M}|} \mathbf{1} & \mbox{for $\pi$-message}, \\ \mathbf{1} & \mbox{for $\lambda$-message}, \end{array} \right . \sigma[\mathbf{m}] = \mathbf{0}, \mu^*[\mathbf{m}] = \mathbf{0}, 
$$
\STATE for each node $Y$ represented in $\mathbf{e}$
set $\mu[\pi_Y(y)] = \mu[\lambda_Y(y)] = \delta_{y,e_y}$ and $\sigma[\pi_Y(y)] = \sigma[\lambda_Y(y)] = 0$ for all $y \in \mathbb{Y}$
\WHILE {$\max \{|\mu[\bm{m}]-\mu^*[\bm{m}]|:\mathbf{m} \in \mathcal{M}^*\}  > \epsilon$}
\STATE $\mu^*[\bm{m}] \leftarrow \mu[\bm{m}]$ for all $m \in \mathcal{M}^*$
\WHILE {$\mathcal{M} \ne \emptyset$} 
\STATE Randomly select $\bm{m}\in \mathcal{M}$, $\mathcal{M} \leftarrow \mathcal{M} \setminus \{\bm{m}\}$ 
\IF {$\bm{m}$ is a $\pi$-message}
\STATE Update $\mu[\bm{m}]$ and $\sigma[\bm{m}]$ via (\ref{pifusion}) and (\ref{spifusion}), respectively
\IF {all $\pi$-messages to node $Z$ have been updated in current round and $Z$ is not in $\mathbf{e}$} \STATE update $\mu[\bm{\pi}_Z]$ and $\sigma[\bm{\pi}_Z]$ via (\ref{fprop}) and (\ref{sfprop}), respectively \ENDIF
\ELSE
\STATE Update $\mu[\bm{m}]$ and $\sigma[\bm{m}]$ via (\ref{backprop}) and (\ref{sbackprop}), respectively
\IF {all $\lambda$-messages to node $X$ have been updated in current round and $X$ is not in $\mathbf{e}$} \STATE update $\mu[\bm{\lambda}_X]$ and $\sigma[\bm{\lambda}_X]$ via (\ref{lambdafusion}) and (\ref{slambdafusion}), respectively \ENDIF
\ENDIF
\ENDWHILE
\STATE $\mathcal{M} \leftarrow \mathcal{M}^*$
\ENDWHILE
\STATE Compute $\mu[\mathbf{p}(Y|\mathbf{e})]$ and $\sigma[\mathbf{p}(Y|\mathbf{e})]$ via (\ref{finalfusion}) and (\ref{sfinalfusion}), respectively, for all nodes in the BN
\end{algorithmic}
\end{algorithm}
\vspace{-10pt}

The initial $\pi$-messages that each node $Y$ sends to each child $Z$ are set so that the mean and covariance matrix of an individual message is
$$
\mu[\bm{\pi}_{Z,Y}] = \frac{1}{|\mathbb{Y}|} \mathbf{1}_{\mathbb{|Y|}}, \hspace{.05in} \mbox{and} \hspace{.05in} \sigma[\bm{\pi}_{Z,Y}] = \mathbf{0},
$$
respectively.  Note that the mean values of the message must sum to one. Also, $\mathbf{1}_N$ is an $N$-length vector of all ones. Similarly, the initial $\lambda$-messages that each node sends to each parent $X$ are set so that the mean and covariance matrix of a single message is
$$
\mu[\bm{\lambda}_{X,Y}] = \mathbf{1}_{\mathbb{|X|}}, \hspace{.05in} \mbox{and} \hspace{.05in} \sigma[\bm{\lambda}_{X,Y}] = \mathbf{0},
$$
respectively.  Unlike the $\pi$-messages, the $\lambda$-messages do not need to be normalized to have their mean values to sum to one. These message values do not have any influence on biasing beliefs to any particular variable value in the execution on the belief propagation operations. The zero covariances also do not influence the variances of the belief. Initially, all the influence of beliefs lies in the mean and covariances of the conditional probabilities and the values of the evidence $\mathbf{e}$.

The update of the message to a parent from a node is accomplished by the back propagation operation in (\ref{backprop}). The means are propagated in the obvious way (i.e., the input messages in (\ref{backprop}) are now the respective means). The covariance for message from node $Y$ to parent $X_i$ is 
\begin{equation} \label{sbackprop}
\begin{aligned}
\sigma[&\lambda_{X_i,Y}(x_i)] = \sum_{\mathbf{X} \sim x_i}  \mathbf{J}_{\theta_{Y|\mathbf{x}}} \sigma[\bm{\theta}_{Y|\mathbf{x}}] \mathbf{J}_{\theta_{Y|\mathbf{x}}}^T\\ & + \sum_{j\ne i} \mathbf{J}_{\pi_{Y,X_j}} \sigma[\bm{\pi}_{Y,X_i}] \mathbf{J}_{\pi_{Y,X_j}}^T
+\mathbf{J}_{\lambda_Y}\sigma[\bm{\lambda}_Y] \mathbf{J}_{\lambda_Y}^T\;,
\end{aligned}
\end{equation}
where the elements of the Jacobians $\mathbf{J}_{\theta_{Y|\mathbf{x}}}$, $\mathbf{J}_{\pi_{Y,X_j}}$, and $\mathbf{J}_{\lambda_Y}$ are given by:
\begin{equation}
\frac{\partial \lambda_{X_i,Y}(x_i)}{\partial \theta_{y|\mathbf{x}}}=\mu[\lambda_Y(y)] \prod_{j\ne i} \mu[\pi_{Y,X_j}(x_j)]  \;,
\end{equation}
\begin{equation}
\begin{aligned}
&\frac{\partial\lambda_{X_i,Y}(x_i)}{\partial \pi_{Y,X_j}(x_j)} = \\ &\sum_{Y}\mu[\lambda_Y(y)]\left(\sum_{\mathbf{x}\sim x_i, x_j} \mu[\theta_{y|\mathbf{x}}]  \prod_{\ell \ne i,j} \mu[\pi_{Y,X_{\ell}}(x_{\ell})]\right)  \;,
\end{aligned}
\end{equation}
and
\begin{equation}
\frac{\partial \lambda_{Y,X_i}(x_i)}{\partial \lambda_Y(y)} = \sum_{\mathbf{X}\sim x_i} \mu[\theta_{y|\mathbf{x}}] \prod_{j \neq i}\mu[\pi_{Y,X_j}(x_j)]\;,
\end{equation}
respectively.

Similarly, the computation for the mean of the message from node $Y$ to its child $Z_j$ is computed by (\ref{pifusion}) in an obvious manner. The covariance is computed as
\begin{equation} \label{spifusion}
\sigma[\bm{\pi}_{Z_i,Y}]= \mathbf{J}_{\pi_Y} \sigma[\bm{\pi}_Y]\mathbf{J}_{\pi_Y} + \sum_{j \ne i} \mathbf{J}_{\lambda_{Y,Z_j}} \sigma[\bm{\lambda}_{Y,Z_j}]\mathbf{J}_{\lambda_{Y,Z_j}}^T,
\end{equation}
where again the Jacobians $\mathbf{J}_{\pi_Y}$ and $\mathbf{J}_{\lambda_{Y,Z_j}}$ are diagonal with non-zero values
\begin{equation}
\frac{\partial \pi_{Z_i,Y}(y)}{\partial \pi_{Y}(y)} = \frac{\mu[\pi_{Z_i,Y}(y)]}{\mu[\pi_{Y}(y)]} \hspace{.1in} \mbox{and} \hspace{.1in}
\frac{\partial \pi_{Z_i,Y}(y)}{\partial \lambda_{Y,Z_j}(y)} = \frac{\mu[\pi_{Z_i,Y}(y)]}{\mu[\lambda_{Y,Z_j}(y)]},
\end{equation}
respectively. 

The update of an internal $\pi$ value is accomplished by the forward propagation in (\ref{fprop}). Again, the means are propagated in the obvious way, and the covariance is updated as
\begin{equation} \label{sfprop}
\begin{aligned}
\sigma[\bm{\pi}_Y]= &\sum_{\bm{X}} 
\mathbf{J}_{\theta_{Y|\mathbf{x}}} \sigma[\bm{\theta}_{Y|x_1,\ldots,x_{N_p}}] \mathbf{J}_{\theta_{Y|\mathbf{x}}}^T +\\ & \sum_{i=1}^{N_p} \mathbf{J}_{\pi_{Y,X_i}} \sigma[\bm{\pi}_{X_i}] \mathbf{J}_{\pi_{Y,X_i}}^T\;,
\end{aligned}
\end{equation}
where the Jacobian
\begin{equation}
\mathbf{J}_{\theta_{Y|\mathbf{x}}} = \left( \prod_{i=1}^{N_p} \mu[\pi_{Y,X_i}(x_i)]\right) \mathbf{I},
\end{equation}
$\mathbf{I}$ is the identity matrix, and
the elements of the Jacobian 
$\mathbf{J}_{\pi_{Y,X_i}}$ are
\begin{equation}
\frac{\partial \pi_Y(y)}{\partial \pi_{Y,X_i}(x_i)} = \sum_{\mathbf{X}\sim x_i} \mu[\theta_{y|\mathbf{x}} ]\prod_{j\ne i} \mu[\pi_{Y,X_j}(x_j)].
\end{equation}

Similarly, the internal $\lambda-$value is updated via (\ref{lambdafusion}) for the mean, and the covariance update is given by
\begin{equation} \label{slambdafusion}
\sigma[\bm{\lambda_Y}] = \sum_{j=1}^{N_c} \mathbf{J}_{\lambda_{Y,Z_j}} \sigma[\bm{\lambda}_{Y,Z_j}]\mathbf{J}_{\lambda_{Y,Z_j}}^T\;,
\end{equation}
where the Jacobians $\mathbf{J}_{\lambda_{V}(X_i)}$ are diagonal with diagonal terms
\begin{equation}
\frac{\partial \lambda_Y(y)}{\partial \lambda_{Y,Z_j}(y)} = \frac{\mu[\lambda_Y(y)]}{\mu[\lambda_{Y,Z_i}(y)]}.
\end{equation}

Once all second-order statistics between the $\pi-$ and $\lambda$ messages between the nodes have been updated along with the internal values, the round is completed. Then, the mean of the queried probabilities for each node is computed via  (\ref{finalfusion}), and the  covariances are given by
\begin{equation} \label{sfinalfusion}
\sigma[p(Y|\mathbf{e})] = \mathbf{J}_{\lambda_Y} \sigma[\bm{\lambda}_Y] \mathbf{J}_{\lambda_Y} + \mathbf{J}_{\pi_Y} \sigma[\bm{\pi}_Y]\mathbf{J}_{\pi_Y}^T
\end{equation}
where the elements of the Jacobians $\mathbf{J}_{\lambda_Y}$ and $\mathbf{J}_{\pi_Y}$ are given by
\begin{equation}
\frac{\partial {p(y|\mathbf{e})}}{\partial \lambda_Y(y')} = \frac{\mu[p(y|\mathbf{e})](\delta_{y,y'}-\mu[p(y'|\mathbf{e})])}{\mu[\lambda_Y(y')]} \hspace{.2in} \mbox{and}
\end{equation}
\begin{equation}
\frac{\partial {p(y|\mathbf{e})}}{\partial \pi_Y(y)} = \frac{\mu[p(y|\mathbf{e})](\delta_{y,y'}-\mu[p(y'|\mathbf{e})])}{\mu[\pi_Y(y')]},  \end{equation}
respectively. 



\section{Numerical Evaluation}
\label{sec:evaluation}

This section provides empirical evidence that SOLBP determines the mean and variance of queries in the form of $p(x_i|\mathbf{e})$ for each unobserved variable that are consistent with those computed via second-order SPN (SOSPN) as described in \cite{cerutti2022handling,kaplan.ivanovska.TBD}. Furthermore, empirical results demonstrate the computational efficiency of SOLBP relative to SOSPN, thus showing the advantages of SOLBP for larger DAG BNs.


In order to make the appropriate comparisons, we set up a test framework in MATLAB which takes BNs as input. During each experimental trial, conditional probabilities, $\bm{\theta}$, of the ground truth BN were individually sampled from a uniform Dirichlet distribution. Then, a sparse but complete set of observations of the network variables were sampled from the appropriate discrete distributions associated with the ground truth BN. These $N_{\textrm{train}}$ examples form the training data for an exact Bayesian learning process.
During each trial, the first step utilizes the $N_{\textrm{train}}$ number of complete observations of the network to compute the parameters of the Dirichlet distribution by counting the number of examples consistent with variable values
(see Section~\ref{sec:UBN}). 

For the inference step, we instantiate a specific set of BN variables as the observed evidence, $\mathbf{e}$, leaving the remaining variables unobserved. 
Then, we infer $p(Y_i|\mathbf{e})$ via SPN over the BN with ground truth conditional probabilities to establish the ground truth inferences. Finally, we infer the mean and variances of the query via SOLBP or SOSPN over the uncertain BN learned via the $N_{\textrm{train}}$ samples. 
These inferred means and variances are recorded over $N_{\textrm{runs}}$ number of trials, where we used $N_{\textrm{runs}}=1000$, and a new ground truth BN was generated after every $100$ trials. Note that, while the ground truth BN remained static for $100$ trials at a time, a new uncertain BN 
is learned on each trial since new Monte Carlo observations of the variables are generated on a per-trial basis.

In order to evaluate accuracy, the recorded means and variances from both the SOSPN and SOLBP methods are plotted against each other. Ideally, we would like to see that the inferences match in all cases, resulting in a diagonal line of slope one. Additionally, we follow the method from \cite{kaplan2018efficient} for generating plots of the desired confidence bound divergence (DeCBoD) for each second-order inference method by establishing confidence bounds from the means and variances. The DeCBoD determines the fractional rate at which the ground truth probabilities fall within 
confidence intervals of significance level $\gamma$ around the inferred probabilities. For $\gamma \in [0,0.99]$, we compute the fractional in-bounds rate and plot it against $\gamma$. 


We tested SOLBP on three sets of BNs, varying types and sizes: (1) three-node networks; (2) small loopy networks; and (3) a large loopy network. Recall, one of the primary motivations for SOLBP is that, similar to the SOSPN method from \cite{cerutti2022handling}, SOLBP can be applied to DAGs with loops, in addition to polytrees. However, we first want to verify that SOLBP can perform equivalently to the SOSPN method on polytrees, so we show the results for both polytrees and DAGs with loops in the following subsections. We provide a discussion on the relative computational efforts in Section \ref{sec:computational-effort}.

\subsection{Three-Node Networks}

\begin{figure}[htp]
    \centering
    \includegraphics[scale=0.25]{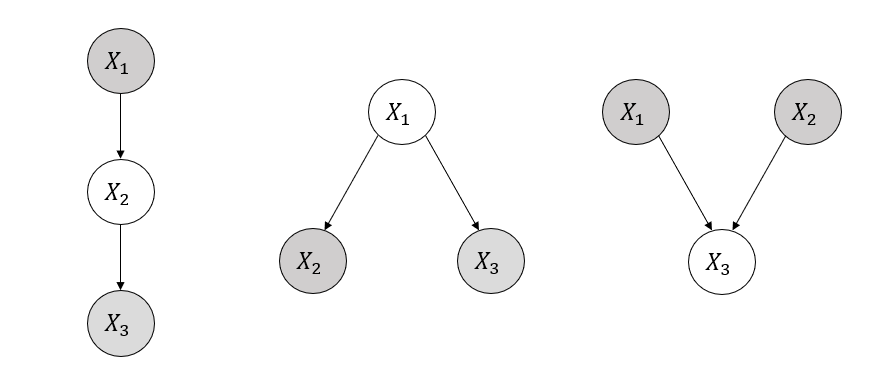}
    \caption{(a) 3-node chain, (b) 3-node tent, (c) 3-node V.}
    \label{fig:threenodenets}
\end{figure}

\begin{figure}[htp]
\subfloat{\includegraphics[width=1.75in]{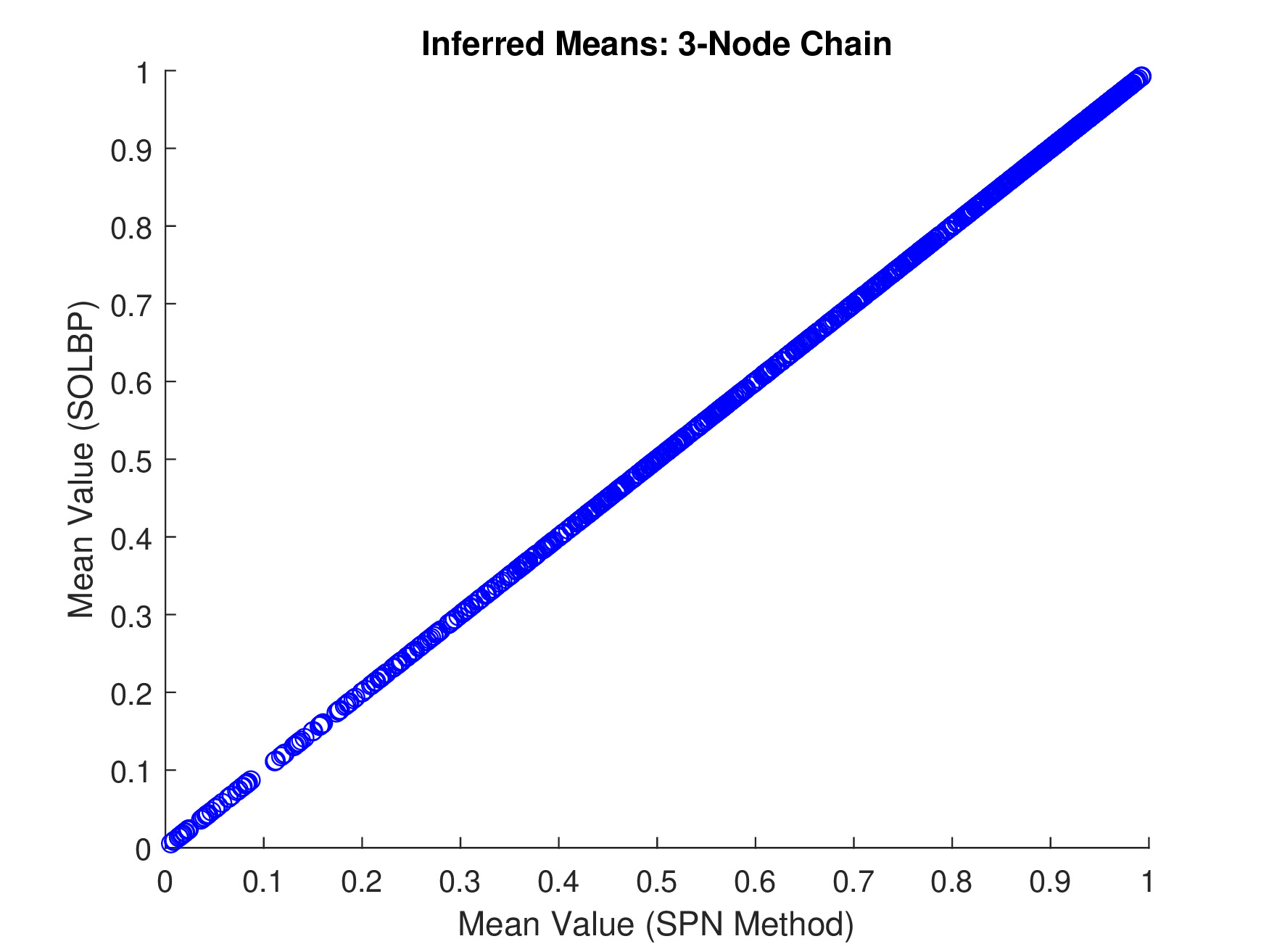}}
\subfloat{\includegraphics[width=1.75in]{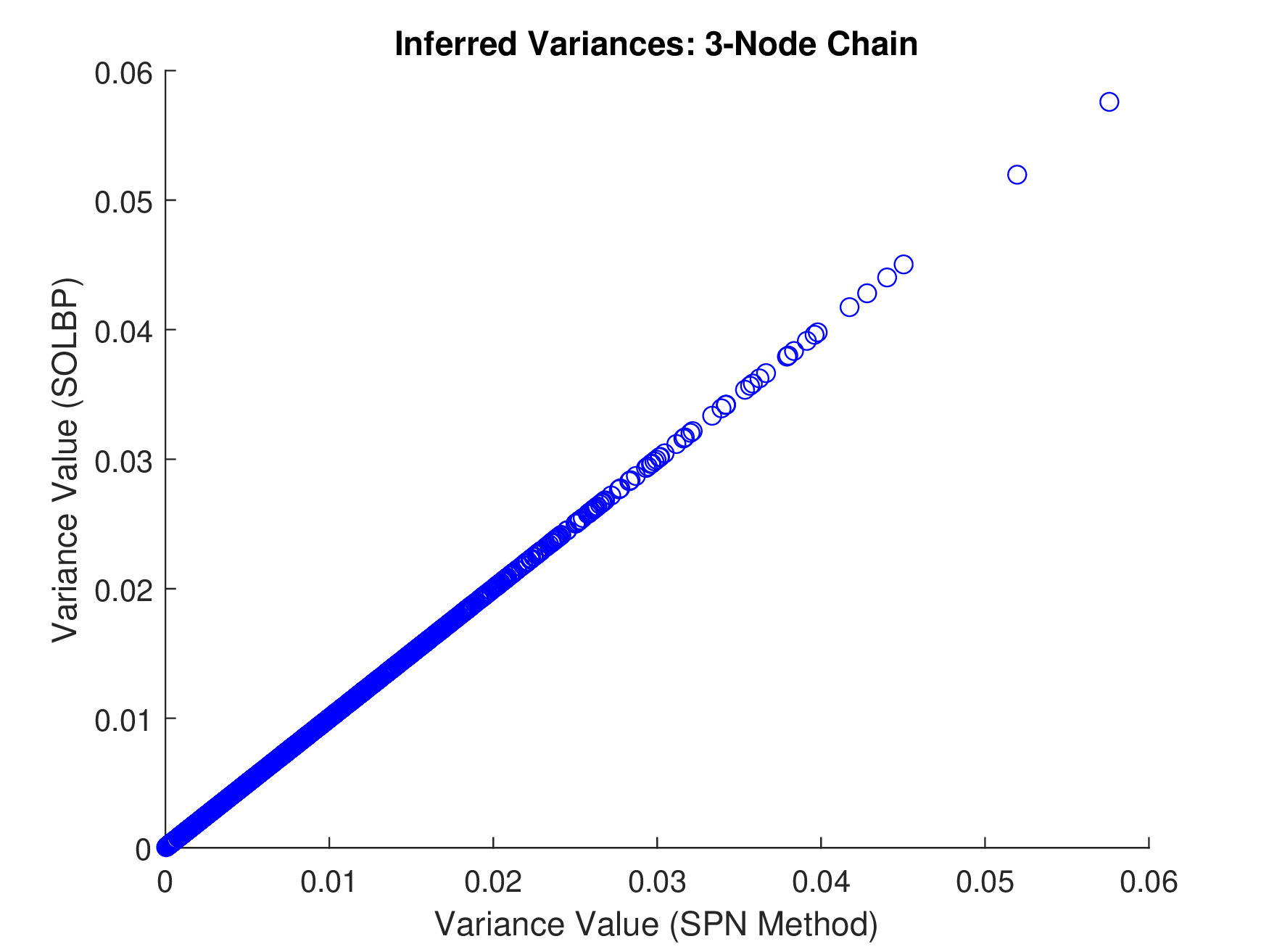}}
\caption{3-node chain inferred means (left) and variances (right).}
\label{fig:3nodechainmeanvar}
\end{figure}

We first tested three simple three-node networks in Figure~\ref{fig:threenodenets}, with binary-valued variables. Values of the means and variances of the inferred probabilities from SOLBP were plotted against the means and variances inferred by the SOSPN method. Evidence variables were chosen as the nodes incident with exactly one edge, regardless of edge direction, which are colored grey in the BN figures for clarity. Figure~\ref{fig:3nodechainmeanvar}, 
compare the inferred means and variances for the three-node chain network over the two methods. The other two networks lead to similar plots.

\vspace{15pt}



\begin{figure}[h]
    \centering
    \includegraphics[width=1.75in]{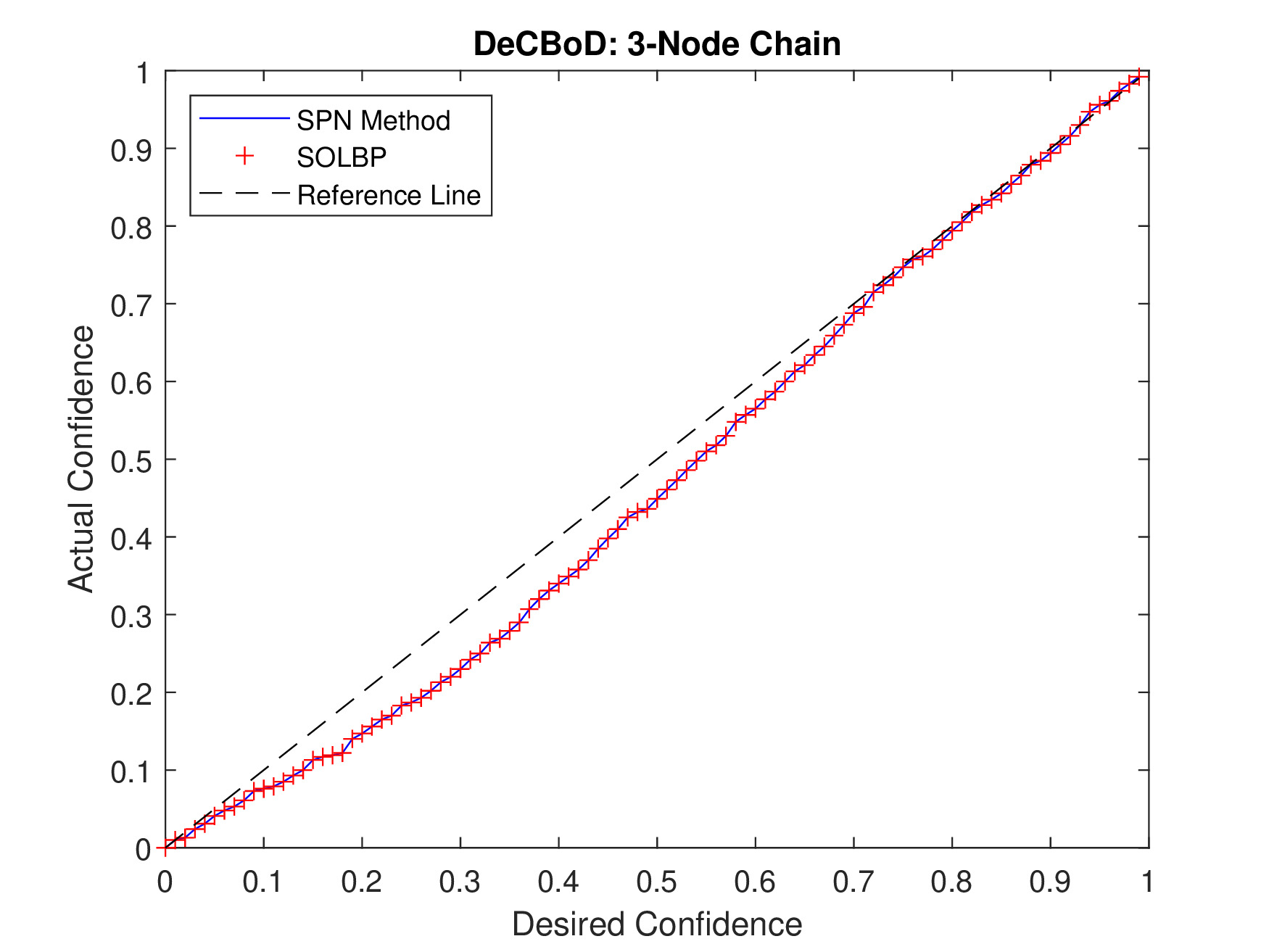}
    \caption{DeCBoD plot for the 3-node chain network.}
    \label{fig:decbod3nodechain}
\end{figure}

The results demonstrate that SOLBP generates inferences which exactly match those generated by the SPN method, up to numerical precision, for these three polytree BNs. Given that the inferred means and variances match, we also expect matching DeCBoD curves. Figure~\ref{fig:decbod3nodechain} shows a sample DeCBoD plot for the three-node chain network from Figure \ref{fig:threenodenets}(a), with the results from SOLBP identified by red crosses and the SPN results identified by the blue lines. DeCBoD plots for the remaining polytrees are similar. The DeCBOD curves show that actual confidence matches well with the desired confidence, meaning that the inferred distributions (mean and variance) is well calibrated to the true uncertainty.

\subsection{Small Loopy Networks}
\begin{figure}[h]
    \centering
    \includegraphics[scale=0.25]{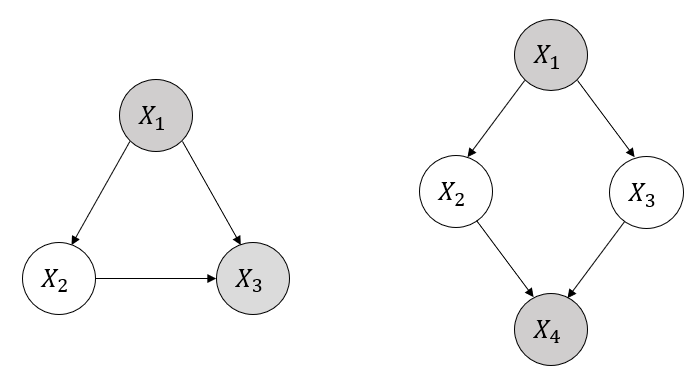}
    \caption{(a) Triangle loopy BN, (b) Diamond loopy BN.}
    \label{fig:loopynets}
\end{figure}

Next, we verified inference accuracy on two basic networks with loops, shown in Figure~\ref{fig:loopynets}. In the case of loopy networks, the number of iterations of SOLBP may affect the resulting inferences, depending on the stopping condition used. Here, we ran SOLBP until the absolute maximum difference between any inferred means from the current and prior iteration were bounded by $\epsilon=1\times 10^{-8}$. 
The observed nodes in the evidence set are shaded grey in Figure \ref{fig:loopynets}. 

\begin{figure}[!ht]
\subfloat{\includegraphics[width=1.75in]{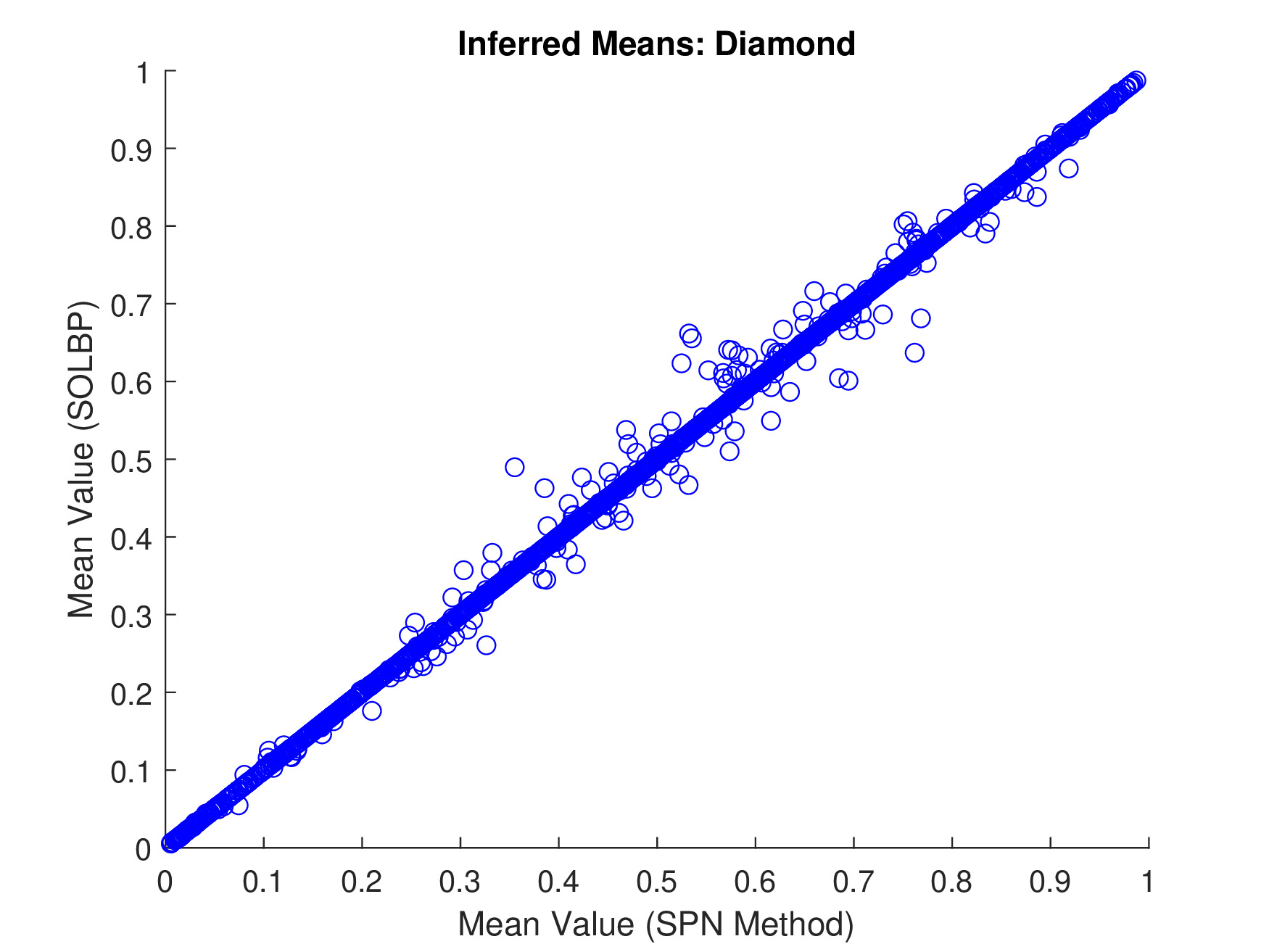}}
\subfloat{\includegraphics[width=1.75in]{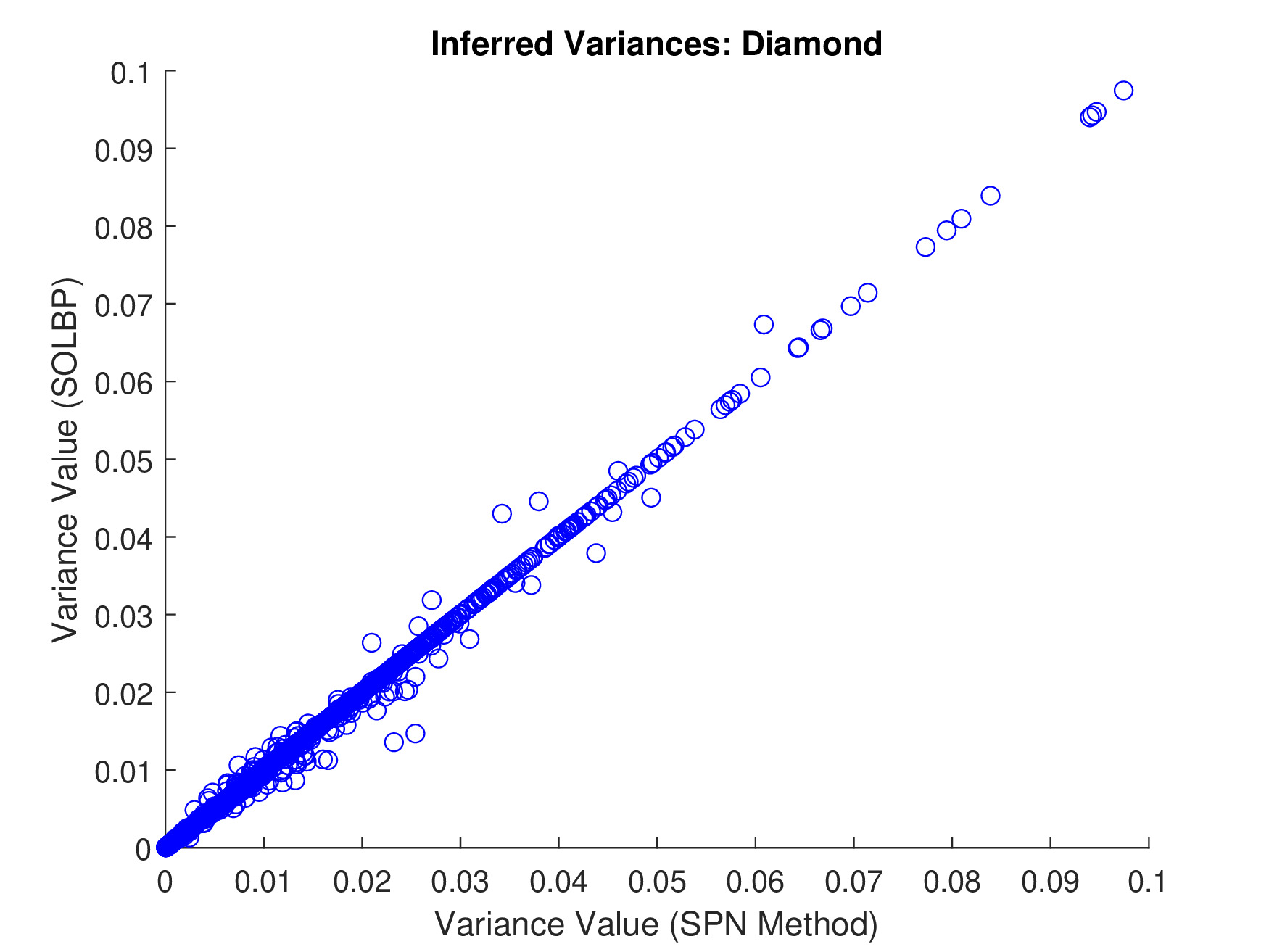}}
\caption{Diamond BN inferred means (left) and variances (right).}
\label{fig:diamondmeanvar}
\end{figure}

\begin{figure}[!ht]
\subfloat{\includegraphics[width=1.75in]{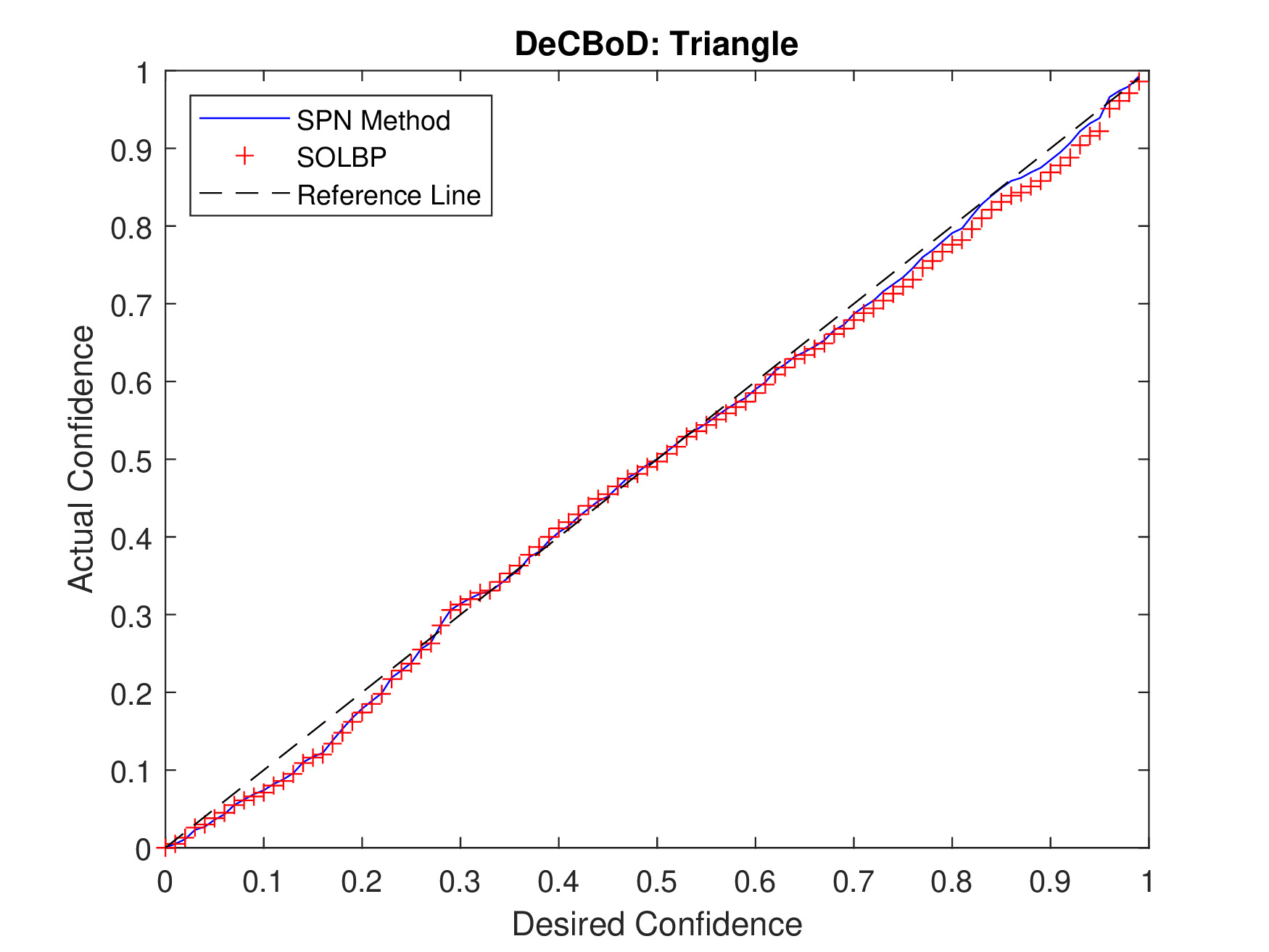}}
\subfloat{\includegraphics[width=1.75in]{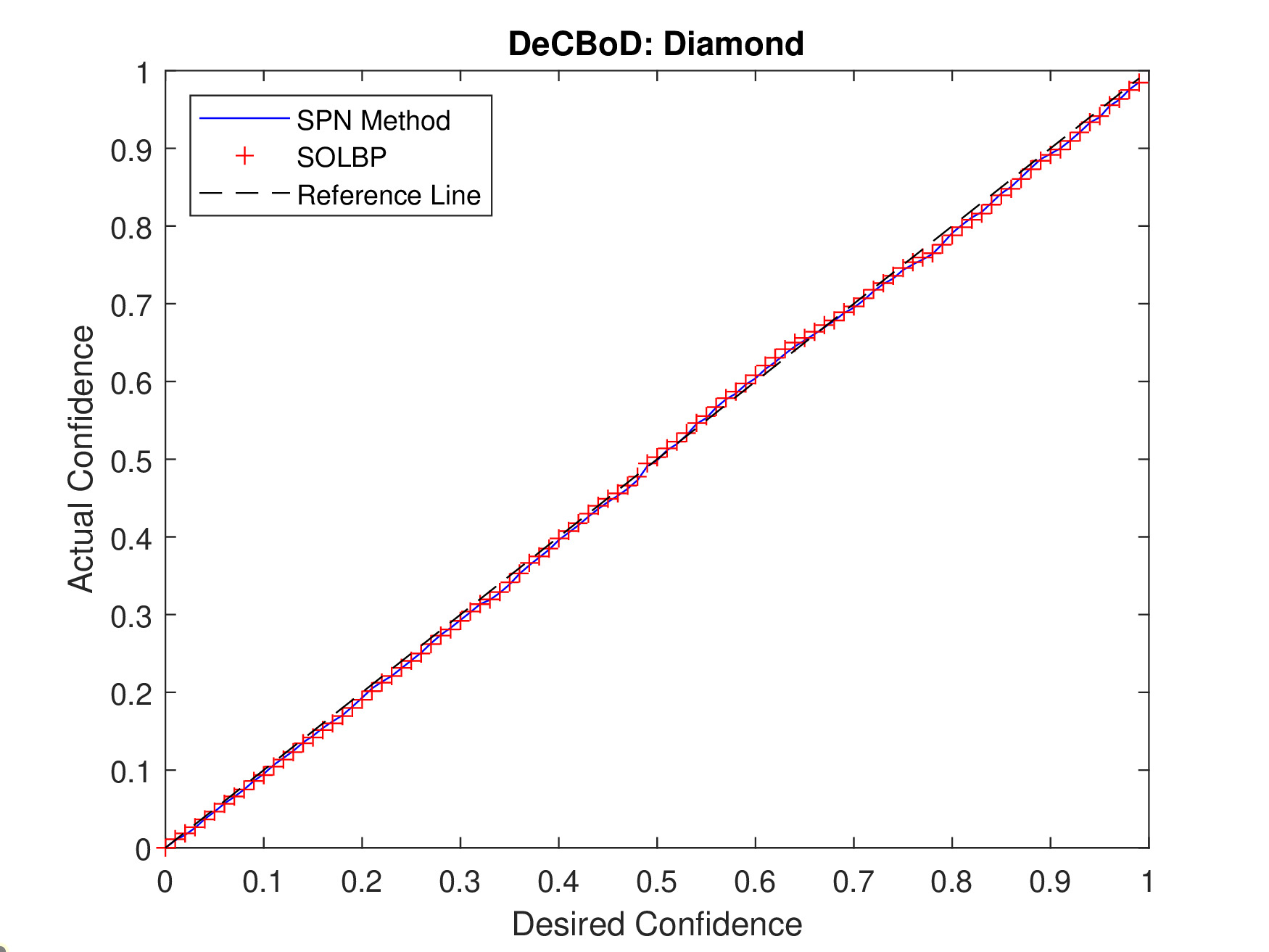}}
\caption{DeCBoD plots for triangle loopy BN (left) and diamond loopy BN (right)}
\label{fig:decbodloopy}
\end{figure}

Figure \ref{fig:diamondmeanvar} compares the resulting inferred means and variances for SOLBP and SOSPN over the diamond network. The triangle network produces similar plots. 
Figure \ref{fig:decbodloopy} shows the DeCBoD plots for the triangle and diamond BNs.

\vspace{15pt}
Figure~\ref{fig:diamondmeanvar} show that the SOLBP and SPN methods produce inferences of means and variances which are similar. 
However, unlike the polytree networks, the values do not all match completely, meaning that some of the scatter points lie off the diagonal. In this particular case with the chosen evidence variables, simply using a more strict stopping condition would produce a complete match, in a finite number of iterations, which we verified. However, for general, multi-loop loopy networks, SOLBP may never converge to the correct values, as is the case with first-order LBP \cite{ihler.05}. On the other hand, for polytrees, within a few runs of message passing, all nodes will have sent and received $\lambda$ and $\pi$-messages with means and covariances consistent with exact inference methods. For single-loop BNs, we would expect eventual convergence of SOLBP over an infinite number of iterations, similar to first-order LBP. Despite the theoretical limitations, empirical evidence demonstrates that the stopping condition we chose for SOLBP produces approximate means and variances that are close to the values generated by the SPN method for small, single-loop networks.

\subsection{A Large Loopy Network}

\begin{figure}[htp]
    \centering
    \includegraphics[scale=0.35]{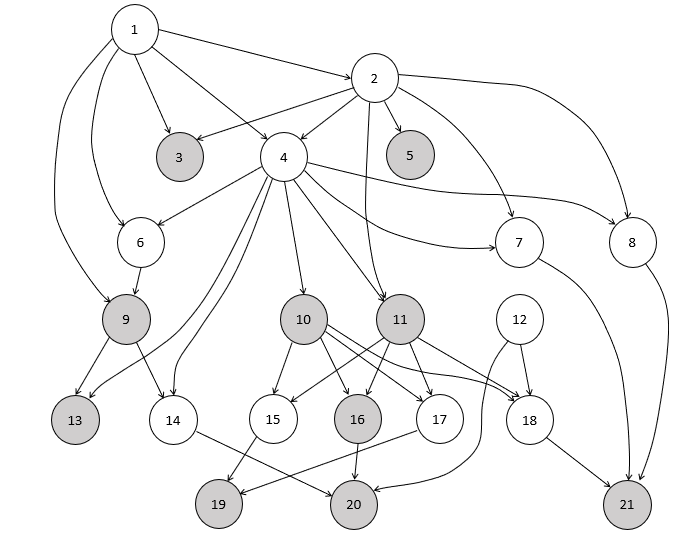}
    \caption{A larger 21-node BN with multiple loops. Grey nodes were used as evidence for the experimental results shown. 
    }
    \label{fig:largenet}
\end{figure}

On a more practical front, we wanted to verify inference approximation accuracy of SOLBP for a larger network with more parameters and multiple loops. Therefore, we created a BN with 21 nodes, as shown in Figure~\ref{fig:largenet}. 
Approximately half of the BN nodes were chosen randomly as evidence variables, which are shaded grey in the figure. While the previous experiments were limited to binary-valued BN variables, we allowed the BN variables to be either binary or tri-valued, to increase the number of total parameters in the BN. We are interested in testing more parameters, both due to potential noise introduced by the increased degrees of freedom as well as to demonstrate potential limitations of either SOLBP or the SOSPN method.

\vspace{15pt}

\begin{figure}[h]
\subfloat{\includegraphics[width=1.75in]{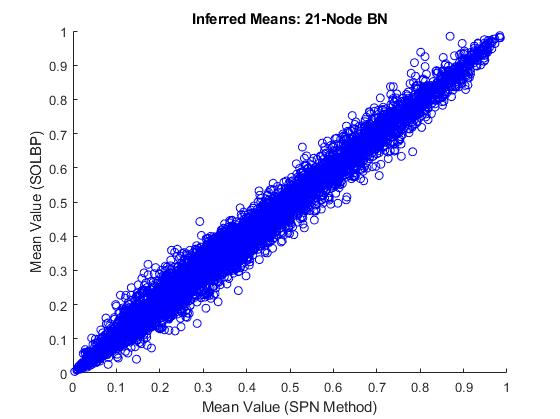}}
\subfloat{\includegraphics[width=1.75in]{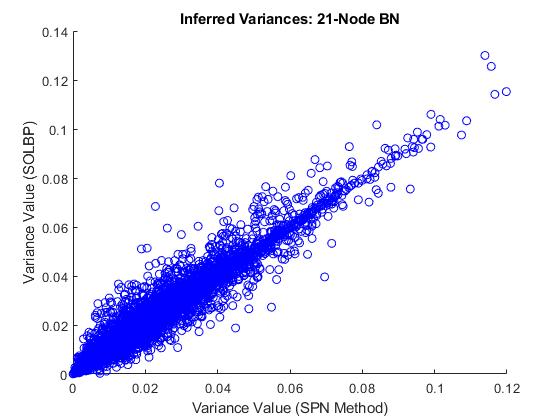}}
\caption{21-node BN inferred means (left) and variances (right).}
\label{fig:21nodemeansvars}
\end{figure}

\begin{figure}[h]
    \centering
    \includegraphics[width=1.75in]{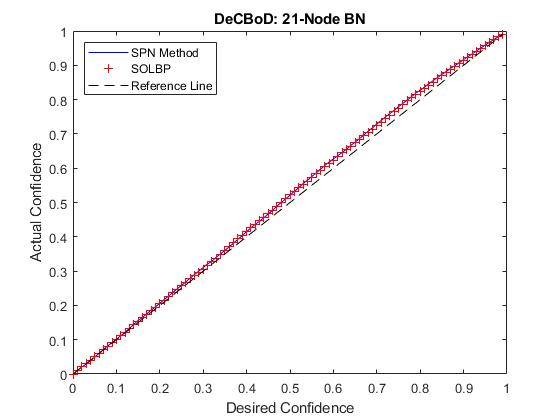}
    \caption{DeCBoD plot for the 21-node network.}
    \label{fig:21nodedecbod}
\end{figure}

Running SOLBP using the same experimental setup on the 21-node network yielded inferred means and variances as shown in Figures~\ref{fig:21nodemeansvars} and~\ref{fig:21nodedecbod}. We see that the SOLBP inferences continue to approximate the inferred values generated by the SOSPN method with relatively high fidelity, though there is a noticeable amount of noise introduced in the larger network setting. As discussed in \cite{murphy1999loopy} and \cite{ihler.05} for first-order LBP, convergence is not guaranteed in BNs with multiple loops, and this may even depend on the values of the conditional probabilities, in addition to the structure. This applies to SOLBP as well. Furthermore, the selection of evidence nodes can impact the rate at which SOLBP converges. It turns out that one can simply decrease the threshold $\epsilon$ used as the stopping condition, or force SOLBP to run for additional iterations, and this will improve the approximation accuracy of SOLBP.

\subsection{Computational Effort}
\label{sec:computational-effort}

Importantly, SOLBP produces inferences that are nearly identical to those produced by the SOSPN method, and SOLBP also scales much better in terms of computational effort. The size of a SPN grows exponentially with respect to the size of the BN as described in \cite{darwiche}.
Table~\ref{tbl:computetime} enumerates run-time metrics indicating that SOLBP is able to execute the second-order inference process multiple times faster than the SOSPN method for moderately sized BNs. This confirms our intuition that SOLBP should be much more scalable than the SPN method.

\vspace{15pt}
\begin{table}[h]
    \centering
    \caption{Comparison of compute time between SOSPN and SOLBP for multiple BN structures.}
    \includegraphics[scale=0.5]{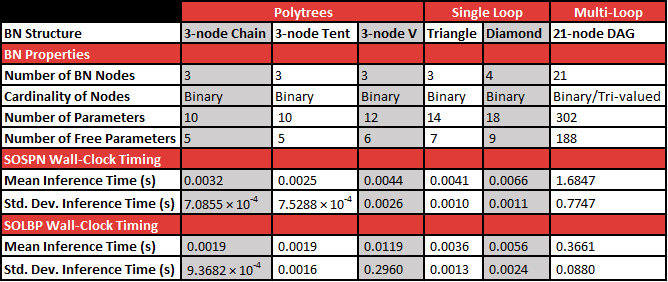}
    \label{tbl:computetime}
\end{table}

\section{Conclusion}
\label{sec:conclusions}
We demonstrated second-order inference using loopy belief propagation. This enables the scalability of second-order belief propagation while in turn allowing for inference over all BNs. Empirical results demonstrate that SOLBP accurately captures the epistemic uncertainty due to limited training data.  The calibration accuracy is comparable to that of SOSPN.  Moreover, SOLBP is signicantly more scalable than SOSPN.



The consistency between SOLBP and SOSPN does increase with more iterations of SOLPB, with a tradeoff of computational efficiency. Fortunately, the DeCBoD plots demonstrate sufficient accuracy of the epistemic uncertainty even for a small number of iterations. Future work will systematically study the accuracy and efficiency of SOLBP over larger and more complex BNs.  Futhermore, we plan to develop theoretical analysis to show why variances from SOLBP should be comparable to that of SOSPN. 



\bibliographystyle{IEEEtran}
\bibliography{ref}

\begin{thebibliography}{10}
\providecommand{\url}[1]{#1}
\csname url@samestyle\endcsname
\providecommand{\newblock}{\relax}
\providecommand{\bibinfo}[2]{#2}
\providecommand{\BIBentrySTDinterwordspacing}{\spaceskip=0pt\relax}
\providecommand{\BIBentryALTinterwordstretchfactor}{4}
\providecommand{\BIBentryALTinterwordspacing}{\spaceskip=\fontdimen2\font plus
\BIBentryALTinterwordstretchfactor\fontdimen3\font minus
  \fontdimen4\font\relax}
\providecommand{\BIBforeignlanguage}[2]{{%
\expandafter\ifx\csname l@#1\endcsname\relax
\typeout{** WARNING: IEEEtran.bst: No hyphenation pattern has been}%
\typeout{** loaded for the language `#1'. Using the pattern for}%
\typeout{** the default language instead.}%
\else
\language=\csname l@#1\endcsname
\fi
#2}}
\providecommand{\BIBdecl}{\relax}
\BIBdecl

\bibitem{pearl.86}
J.~Pearl, ``Fusion, propagation, and structuring in belief networks,''
  \emph{Artificial Intelligence}, vol.~29, no.~3, pp. 241--288, Sep. 1986.

\bibitem{credal98}
M.~Zaffalon and E.~Fagiuoli, ``{2U}: An exact interval propagation algorithm
  for polytrees with binary variables,'' \emph{Artificial Intelligence}, vol.
  106, no.~1, pp. 77--107, 1998.

\bibitem{shenoy.89}
P.~P. Shenoy, ``A valuation-based language for expert systems,'' \emph{Int.
  Journal of Approximate Reasoning}, vol.~3, no.~2, pp. 383--411, 1989.

\bibitem{ivanovska.15}
M.~Ivanovska, A.~J{\o}sang, L.~Kaplan, and F.~Sambo, ``Subjective networks:
  Perspectives and challenges,'' in \emph{Proc. of the 4th International
  Workshop on Graph Structures for Knowledge Representation and Reasoning},
  Buenos Aires, Argentina, Jul. 2015, pp. 107--124.

\bibitem{kaplan2018efficient}
L.~Kaplan and M.~Ivanovska, ``Efficient belief propagation in second-order
  {Bayesian} networks for singly-connected graphs,'' \emph{International
  Journal of Approximate Reasoning}, vol.~93, pp. 132--152, 2018.

\bibitem{rohmer2020uncertainties}
J.~Rohmer, ``Uncertainties in conditional probability tables of discrete
  {Bayesian} belief networks: A comprehensive review,'' \emph{Engineering
  Applications of Artificial Intelligence}, vol.~88, p. 103384, 2020.

\bibitem{cerutti2022handling}
F.~Cerutti, L.~M. Kaplan, A.~Kimmig, and M.~{\c{S}}ensoy, ``Handling epistemic
  and aleatory uncertainties in probabilistic circuits,'' \emph{Machine
  Learning}, pp. 1--43, 2022.

\bibitem{kaplan.ivanovska.TBD}
L.~Kaplan, M.~Ivanovska, K.~V. Mishra, and F.~Cerutti, ``Inference in
  second-order {Bayesian} networks,'' in preparation.

\bibitem{kleiter.96}
G.~D. Kleiter, ``Propagating imprecise probabilities in {Bayesian} networks,''
  \emph{Artificial Intelligence}, vol.~88, no. 1-2, pp. 143--161, 1996.

\bibitem{VanAllen.08}
T.~Van~Allen, A.~Singh, R.~Greiner, and P.~Hooper, ``Quantifying the
  uncertainty of a belief net response: {Bayesian} error-bars for belief net
  inference,'' \emph{Artificial Intelligence}, vol. 172, no. 4-5, pp. 483--513,
  2008.

\bibitem{murphy1999loopy}
K.~P. Murphy, Y.~Weiss, and M.~I. Jordan, ``Loopy belief propagation for
  approximate inference: An emprical study,'' in \emph{Proc. of the 15th Conf.
  on Uncertainty in Artificial Intelligence}, 1999, pp. 467--475.

\bibitem{ihler.05}
A.~T. Ihler, J.~W. Fisher~III, and A.~S. Willsky, ``Loopy belief propagation:
  Convergence and effects of message errors,'' \emph{Machine Learning
  Research}, vol.~6, no.~31, pp. 905--936, 2005.

\bibitem{jensen.07}
F.~V. Jensen and T.~D. Nielsen, \emph{{Bayesian} Networks and Decision
  Graphs}.\hskip 1em plus 0.5em minus 0.4em\relax New York: Springer, 2007.

\bibitem{madsen1999lazy}
A.~L. Madsen and F.~V. Jensen, ``Lazy propagation: A junction tree inference
  algorithm based on lazy evaluation,'' \emph{Artificial Intelligence}, vol.
  113, no. 1-2, pp. 203--245, 1999.

\bibitem{darwiche}
A.~Darwiche, ``A differential approach to inference in {Bayesian} networks,''
  \emph{Journal of the ACM}, vol.~50, no.~3, pp. 280--305, 2003.

\end{thebibliography}

\end{document}